\documentclass[conference]{IEEEtran}

\input epsf
\usepackage{graphicx}
\usepackage{amsmath}
\usepackage{hyperref}
\usepackage{algorithm}
\usepackage{array}
\usepackage{amsfonts}
\usepackage{amssymb}
\usepackage{caption}
\usepackage{booktabs}
\usepackage{multirow}
\usepackage{algpseudocode}
\usepackage{tikz}
\usepackage{authblk}
\usepackage{placeins}

\begin{document}

\title{\LARGE DSReLU: A Novel Dynamic Slope Function for Superior Model Training}
\author[1, a]{Archisman Chakraborti}

\author[2]{Bidyut B Chaudhuri}

\affil[1]{Harish Chandra Research Institute, Prayagraj}
\affil[2]{Techno India University, Kolkata, India}
\affil[a]{Corresponding author email:  archismanchakraborti@hri.res.in}


\maketitle

\begin{abstract}
This study introduces a novel activation function, characterized by a dynamic slope that adjusts throughout the training process, aimed at enhancing adaptability and performance in deep neural networks for computer vision tasks. The rationale behind this approach is to overcome limitations associated with traditional activation functions, such as ReLU, by providing a more flexible mechanism that can adapt to different stages of the learning process. Evaluated on the Mini-ImageNet, CIFAR-100, and MIT-BIH datasets, our method demonstrated  improvements in classification metrics and generalization capabilities. These results suggest that our dynamic slope activation function could offer a  new tool for improving the performance of deep learning models in various image recognition tasks.

\end{abstract}
\IEEEoverridecommandlockouts

%
\IEEEpeerreviewmaketitle

\section*{Introduction}
Activation functions are crucial in neural networks as they introduce nonlinearity, enabling the modeling of complex patterns. The Sigmoid function was one of the earliest, mapping inputs to a range between 0 and 1, useful for binary classification. However, it suffers from the vanishing gradient problem and is not zero-centered, complicating optimization \cite{clevert2015elu, ramachandran2017searching}. The Tanh function improved upon this by mapping inputs between -1 and 1, centering data and aiding convergence, yet still faces vanishing gradients and higher computational costs \cite{activation2021systematic}. ReLU, which outputs the input if positive and zero otherwise, mitigates vanishing gradients and accelerates training, though it can lead to "dying ReLU" neurons that become inactive \cite{misra2019mish}. Initially introduced by Yann LeCun and colleagues in the late 1990s, the activation function didn't gain immediate traction. It wasn't until 2010, when Vinod Nair and Geoffrey Hinton used it successfully in deep neural networks for the ImageNet competition, that ReLU started to gain widespread recognition and adoption in the deep learning community \cite{nair2010relu, lecun2012efficient}. ELU introduces an exponential component for negative values to address the vanishing gradient problem but is computationally expensive \cite{clevert2015elu}. Swish, discovered through automated search, interpolates smoothly between linear and nonlinear behaviors, outperforming ReLU in some tasks but also being computationally intensive \cite{ramachandran2017searching}. \\
Our proposed activation function, DSReLU, introduces a dynamic slope that adjusts throughout the training process. DSReLU effectively balances initial steep slopes for rapid learning and gradually decreases them to promote stability as training progresses. \\
We introduce Dynamic Slope Changing Rectified Linear Unit (DSReLU), a novel dynamic slope activation function. DSReLU adapts its slope throughout the training process, enhancing the network's learning efficiency and generalization capabilities by addressing issues such as vanishing gradients and overfitting. \\
The main \textbf{contributions} of this paper are as follows: 
\begin{itemize}
    \item DSReLU achieves superior results compared to state-of-the-art activation functions in classification tasks, in validation metrics such as accuracy, F1-score, and AUC on the \textbf{(1)} Mini-ImageNet
    \textbf{(2)} CIFAR-100
    \textbf{(3)} MIT-BIH datasets.
    \item Our proposed activation function does not get saturated with training data, maintaining effective learning and avoiding the issues of dead neurons.
    \item DSReLU shows enhanced generalization capabilities on new test data.
    \item Our proposed activation function reduces over-fitting during training on its own without the use of external regularisation techniques (like batch normalisation, drop-out, parameter tying, etc.)
\end{itemize}
The rest of the paper is organised as follows. Section 2 reviews the related work on activation functions, discussing their evolution and the motivation for developing new functions. Section 3 details the design and mathematical formulation of DSReLU. Section 4 describes the experimental setup. In Section 5, we present the results and discussion, comparing the performance of DSReLU with other activation functions. Section 6 concludes the paper, summarizing the contributions and suggesting directions for future research. 

\section*{Related Works}
The exploration of novel activation functions seeks to surpass the limitations of traditional ReLU and its variants, leading to significant advancements such as the Leaky ReLU. Leaky ReLU allows a small, non-zero gradient when the unit is inactive (negative side), helping to mitigate the "dying ReLU" problem \cite{maas2013rectifier}. Additionally, the development of learning-based adaptive activation functions, such as Adaptive Piecewise Linear (APL) and Swish, introduces a paradigm shift towards dynamically adaptable neural networks
\cite{activationFunctionsOverview, activationFunctionsSurvey}. These functions feature learnable parameters that adjust the activation response during training, enhancing model adaptability and efficiency \cite{apl, swish}. \\
Mish  \cite{misra2019mish} is a novel self-regularized non-monotonic activation function introduced by Diganta Misra in 2019. It is defined as \( f(x) = x \cdot \tanh(\text{softplus}(x)) \), where \(\text{softplus}(x) = \log(1 + e^x) \). Mish has shown superior performance over traditional activation functions such as ReLU, Leaky ReLU, and Swish across various deep learning tasks. The smooth and continuous nature of Mish helps moderate issues like the dying ReLU problem and improves the flow of gradients during backpropagation, resulting in better generalization and faster convergence. Mish has been effectively utilized in image classification, object detection, and generative modeling, contributing to notable improvements in accuracy and robustness.

The introduction of a dynamic slope-changing Leaky ReLU variant in this work aligns with the evolving trend towards more flexible deep learning models. By proposing an activation function that adapts its behavior based on the training phase, this research contributes to the ongoing efforts to optimize neural network performance across diverse tasks and datasets.
\section{Methodology}

\subsection{Activation Function Design}

The core innovation of this study lies in the development of a new variant of the Rectified Linear Unit (ReLU) activation function, characterized by its dynamically changing slope. This adaptation is predicated on enhancing the responsiveness of the activation function to the progression of training epochs, thereby addressing common challenges in deep learning such as the vanishing gradient problem and the adaptability of the model as it learns.

\subsubsection{Mathematical Formulation:}

The \textit{DSReLU} activation function, denoted as \( f(x; t) \), is designed for neural networks to adaptively modify its behavior during training. Mathematically, it is defined as:

\[
f(x; t) = 
\begin{cases} 
  x \cdot s(t) & \text{if } x > 0, \\
  x & \text{if } x \leq 0,
\end{cases}
\]

where \( s(t) \) represents the dynamic slope for positive input values and is calculated using the expression:

\[
s(t) = a + \frac{b - a}{1 + e^{-k(t - 0.5)}},
\]

with $a$ and $b$ being the initial and final slopes, respectively, and \( k \) being the steepness parameter. The parameter \( t \) indicates the training progress, evolving over time to facilitate the transition from \( a \) to \( b \).

\subsubsection{Rationale:}

The `DSReLU` function was designed to enhance neural network training by dynamically adjusting the activation slope, thereby addressing several challenges, including the vanishing gradient problem. Initially, a steep slope facilitates rapid error signal propagation, enabling swift learning and feature capture. As training progresses, the function's slope gradually decreases, promoting stability and reducing overfitting risk, which is crucial for generalizing to unseen data.

\begin{figure}
\centering
\includegraphics[width=0.3\textwidth]{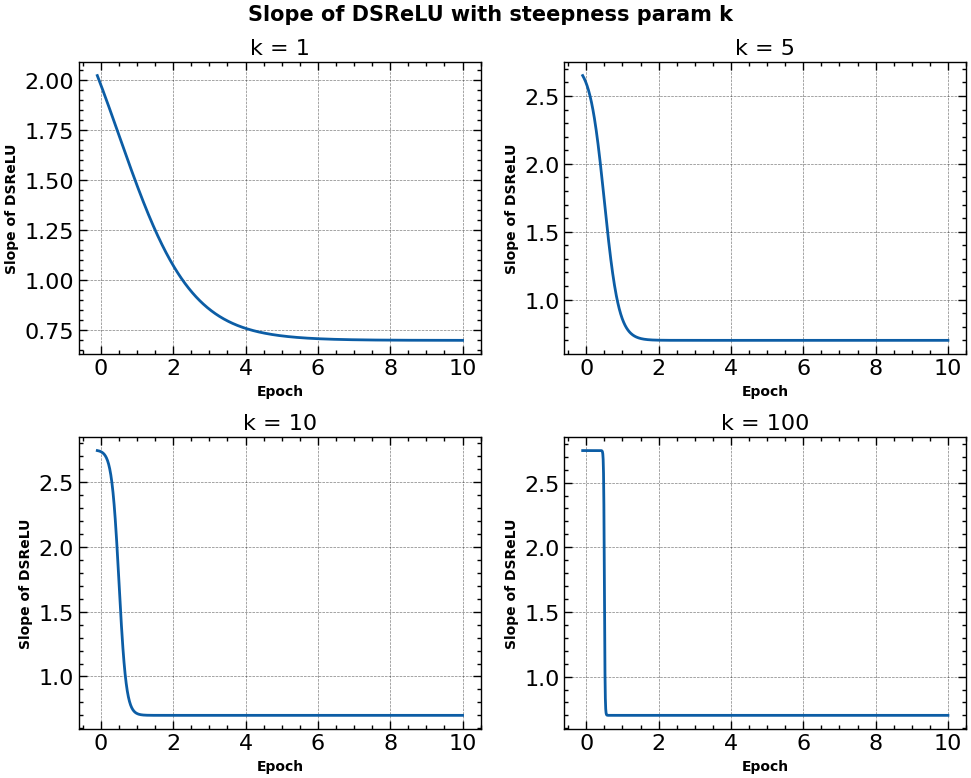}
\caption{Slope change of DSReLU with parameter $k$}\label{fig:dsrelu}
\end{figure}

The steepness parameter $k$ was chosen to be equal to 5. As seen in Fig \ref{fig:dsrelu}, $k$ controls how fast the slope of DSReLU changes from $a$ to $b$. It is the only important data dependent hyperparameter in the whole formulation. A very high value of $k$ leads to a sharp change in the slope resulting in unstable training. A very small value of $k$ results in too slow a change in the slope of DSReLU leading to poor performance. The choice of $k$ was further validated through 5 fold cross-validation.

The parameter $b$ represents the final slope of the slope function $s(t)$ in DSReLU and is chosen to be the tangent of a small angle, specifically the tangent of 10 degrees in this context. This selection is based on the concept that, by the end of training, the gradient of the activation function should stabilize to a minimal value. Similarly, the parameter $a$ indicates the initial slope value of the slope function $s(t)$ in DSReLU. Its value is intended to be as large as possible to maximize the gradient of the activation function during the early stages of training. Therefore, it was set to the tangent of 85 degrees. It was not set to the tangent of 90 degrees due to the numerical instability caused by the tangent function's divergence at 90 degrees.  The choice of $a$ and $b$ can have a degree of flexibility as long as they are confined to the first quadrant.

\subsection{Analysis of DSReLU Function}

The \textit{DSReLU} (Dynamic Smooth ReLU) activation function is designed to optimize neural network training by adapting its properties over time. Its mathematical characteristics are detailed as follows:

\begin{itemize}
    \item \textbf{Monotonicity:} The function is defined as monotonically increasing, satisfying the condition \( f(x_1; t) \leq f(x_2; t) \) for any \( x_1 < x_2 \), ensuring consistent gradient direction which is crucial for stable optimization.

    \item \textbf{Continuity:} \textit{DSReLU} is continuous over \(\mathbb{R}\), providing a seamless function graph that ensures uninterrupted gradient flow, essential for effective gradient descent optimization.

    \item \textbf{Differentiability:} It is piecewise linear for all \( x \in \mathbb{R} \setminus \{0\} \), facilitating reliable gradient computation in backpropagation. At \( x = 0 \), numerical approximations are employed to estimate the gradient, thus maintaining learning continuity. This, however, creates non-linearity which is good for solving/ piece-wise linear problems.

    \item \textbf{Approximation to Identity Near Origin:} Near \( x = 0 \), \textit{DSReLU} approximates the identity function when the initial slope \( a \) is close to one, enabling efficient learning of linear relationships in the initial training phases.

    \item \textbf{Mitigating the Vanishing Gradient Problem:} By ensuring a non-zero gradient for negative values, \textit{DSReLU} prevents the vanishing gradient issue common in standard ReLU activations, thereby keeping neurons active and learning throughout the training process.
\end{itemize}

\section{Experimental Setup and Model Architecture}

\subsection{Model Architecture}
In this study, we utilized the ResNet-34 architecture directly from the PyTorch library \cite{pytorch_resnet34}. ResNet-34, a variant of the residual networks, is known for its depth and ability to attenuate the vanishing gradient problem through the use of residual connections. This architecture, consisting of 34 layers, enables efficient training of deep neural networks by facilitating the flow of gradients during back-propagation. We did not use any regularisation technique in our implementation and just kept the base architecture to compare the performance of the activation functions in question only.

\subsection{Loss Function and Optimizer}

The Cross Entropy Loss \cite{mao2023crossentropy} function was employed as the loss function.
The Adam optimizer \cite{kingma2017adam} was used. 
Constants taken were:
(1)\; Learning rate $\alpha$: $10^{-4}$, (2)\; $\beta_1, \beta_2 \in [0, 1)$: Exponential decay rates for the moment estimates ($\beta_1 = 0.9$ and $\beta_2 = 0.999$),   
(3)\; $\epsilon$: A small scalar to prevent division by zero (= $10^{-8}$) for the Adam optimizer.

\subsection{Hardware Used}
The computational experiments conducted in this study were performed using an Intel i5 12th generation processor coupled with 16 GB RAM. The intensive graphical computations were facilitated by an NVIDIA RTX 3080 graphics card, with a 12 GB RAM.The entire training process was orchestrated using PyTorch\cite{paszke2019pytorchimperativestylehighperformance}.

\subsection{Evaluation Metrics}
To rigorously evaluate the performance of the neural network employing the proposed dynamic slope-changing activation function, we utilized \textbf{Accuracy}, \textbf{F1-score}, \textbf{AUC(Area under the curve) score}.
We used 5-fold cross-validation and presented the results of all five phases in tables to demonstrate robust and reliable performance evaluation.

\section{Results and Discussion}

All tables have been included in the Appendix to save space.

\subsection{Activation Functions Compared}
In our study, we compared the 
\textbf{ReLU (Rectified Linear Unit)}, 
\textbf{Leaky ReLU} (with leak $\alpha = 0.01$),
\textbf{Sigmoid}, \textbf{Tanh (Hyperbolic Tangent)}, \textbf{Mish} with our activation function \textbf{DSReLU}.

\subsection{Mini ImageNet}
The Mini-ImageNet\cite{vinyals2016matching} is often used as a benchmark dataset. It consists of a subset of the larger ImageNet dataset, containing 100 classes with 600 images per class, resulting in a total of 60,000 images. Each image is of size 84x84 pixels.
We used this dataset for our purpose.

\subsubsection{Performance on Training and validation sets during training on Mini Imagenet: }

\begin{figure}
\centering
\includegraphics[width=0.5\textwidth]{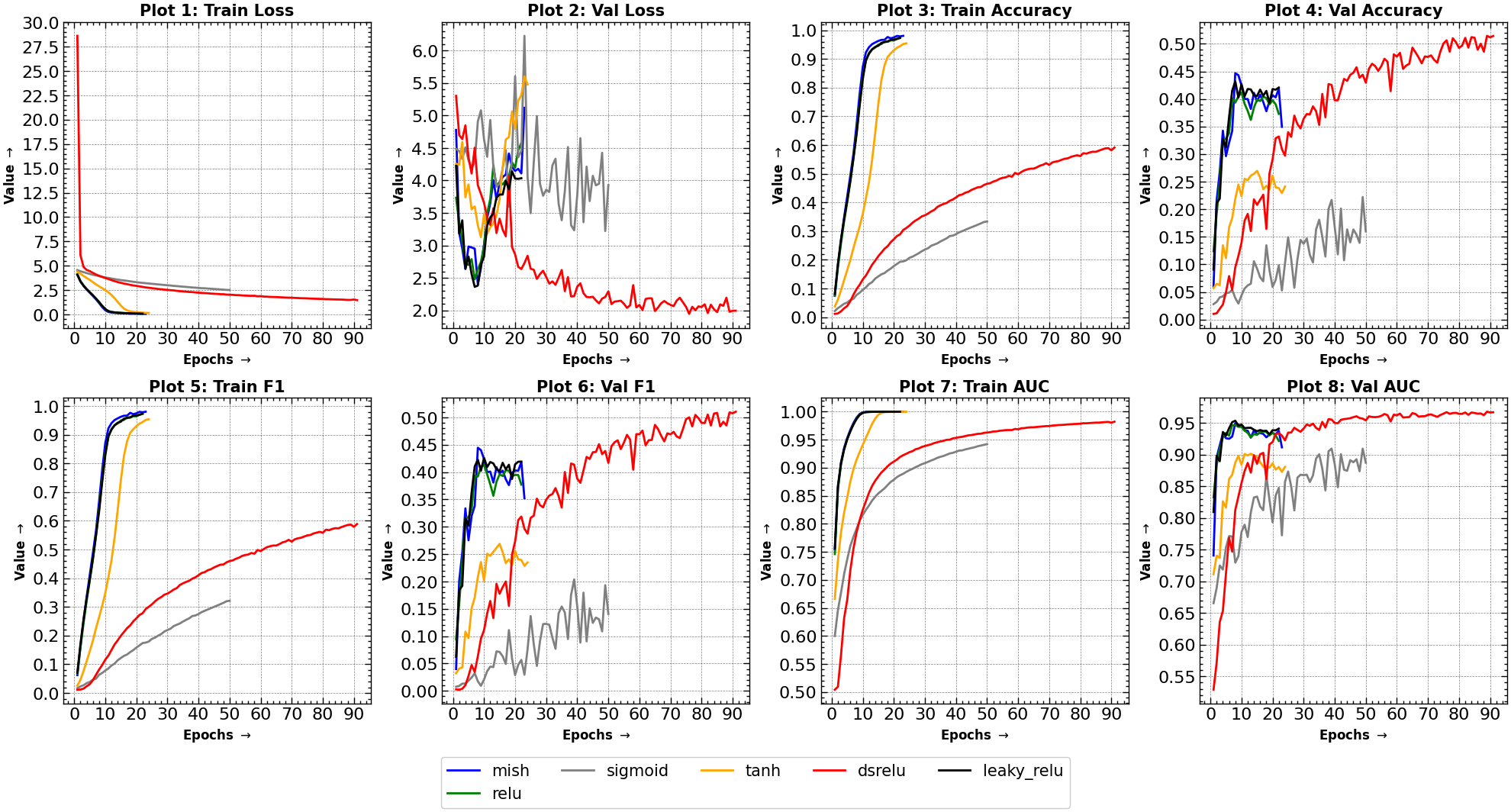}
\caption{Performance of different activation functions on the Mini-ImageNet Dataset}\label{fig:imagenet-metrics}
\end{figure}

Refer to Fig \ref{fig:imagenet-metrics} and Table \ref{Imagenet-table} for details.  
As can be seen, DSReLU lessens the problem overfitting and provides about $20\%$ better validation accuracy and about $20\%$ better F1-score on the validation sets as compared to state-of-the-art activation functions. These figures show the improvement of DSReLU over the second-best performing activation function on the validation set, calculated by subtracting their best validation metrics from DSReLU's (e.g., DSReLU achieves 0.51444 in validation accuracy versus Mish's 0.418778 which leads to an almost 20\% improvement).
 In this context, the term "improvement" means the following:
\begin{equation}
\left( \frac{A_{\text{DSReLU}} - A_{\text{other}}}{A_{\text{other}}} \right) \times 100 \%
\label{improvement}
\end{equation}

\begin{align*}
A_{\text{DSReLU}} & = \text{Best Validation Accuracy of DSReLU}
\end{align*}
\vspace{-0.3in}
\begin{align*}
A_{\text{other}} & = \text{Best Validation Accuracy of the other top-performing
activation}
\end{align*}

\subsection{The CIFAR-100 dataset}

For our experiments, we utilized the CIFAR-100 dataset \cite{krizhevsky2009learning}, which comprises 60,000 32x32 color images distributed across 100 classes, with each class containing 600 images. We used a batch-size of 32 for the training process.

\subsubsection{Performance on Training and Validation sets during training on CIFAR-100} 
\begin{figure}
\centering
\includegraphics[width=0.5\textwidth]{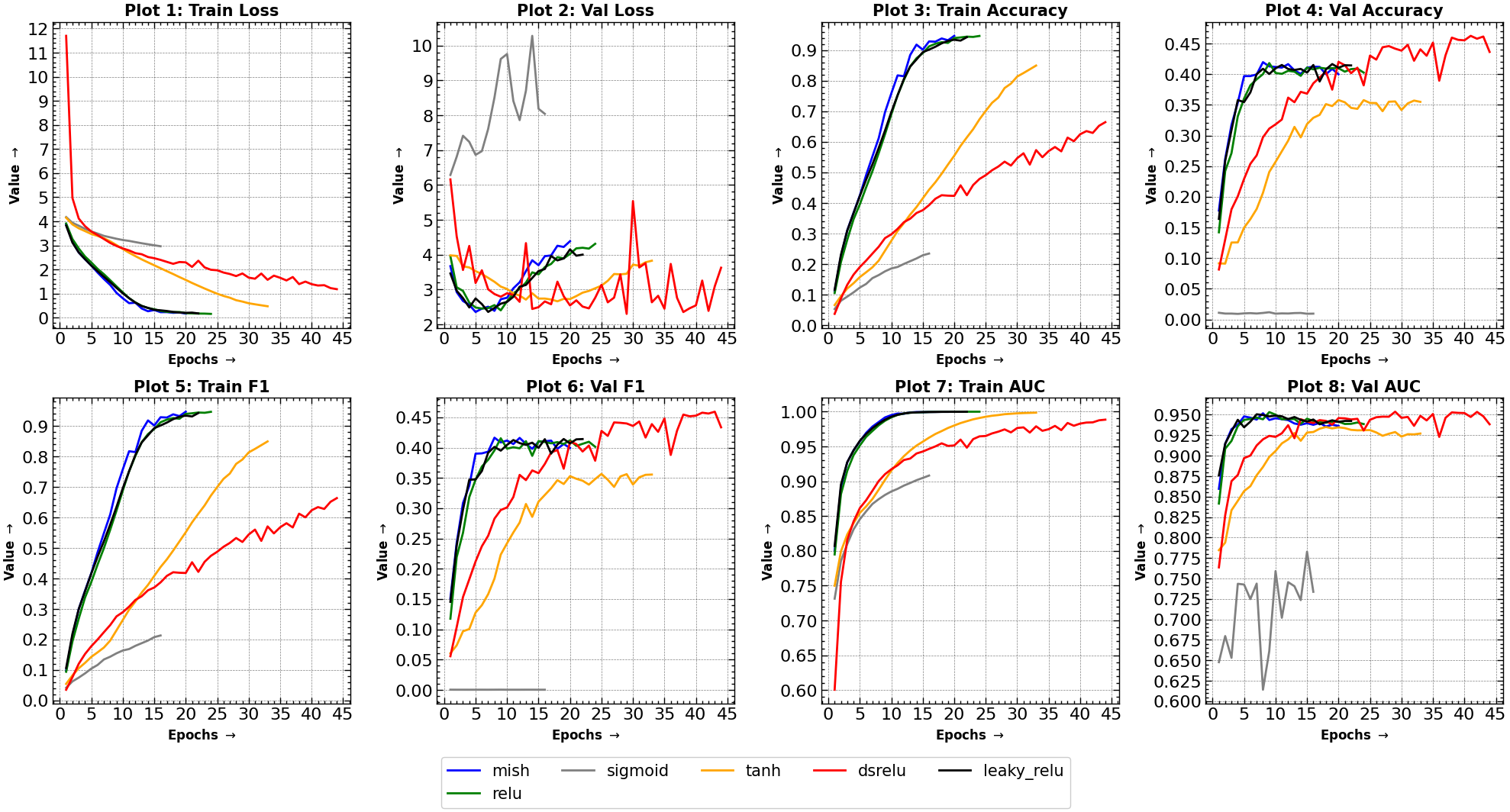}
\caption{Performance of different activation functions on the CIFAR100 Dataset}\label{fig:cifar_training}
\end{figure}

Refer to Fig \ref{fig:cifar_training} and Table \ref{Cifar-table} for details. 
As can be seen from the training and validation plots and the performance metrics, DSReLU abates the problem overfitting and provides about $10\%$ better validation accuracy and about $15\%$ better F1-score on the validation sets as compared to state-of-the-art activation functions. These figures were calculated by subtracting 
the best validation metrics of the second-best performing activation metric on validation set from those of DSReLU (Example: On Validation accuracy, DSReLU gets 0.461300 and LeakyReLU gets 0.416400, which leads to an improvement of about 10\%). The term improvement means the same as in Eqn. \ref{improvement}.

\FloatBarrier

\subsection{MIT-BIH Dataset}
The MIT-BIH Arrhythmia Database\cite{moody2001impact}, comprising 48 half-hour excerpts from two-channel ambulatory electrocardiogram (ECG) recordings, it represents a diverse population of 47 subjects. 
Recordings are digitized at 360 samples per second per channel with an 11-bit resolution over a 10 mV range. Each record typically features two 30-minute ECG lead signals, primarily in MLII and one of V1, V2, or V5 leads. The dataset is annotated, containing over 110,000 beats classified into different arrhythmia categories.

In this analysis, no preprocessing techniques were applied to ease the prevelant class imbalance, potentially affecting the model's performance on underrepresented classes.

\subsubsection{Performance on Training and Validation sets during training on CIFAR-100: } 
\begin{figure}
\centering
\includegraphics[width=0.5\textwidth]{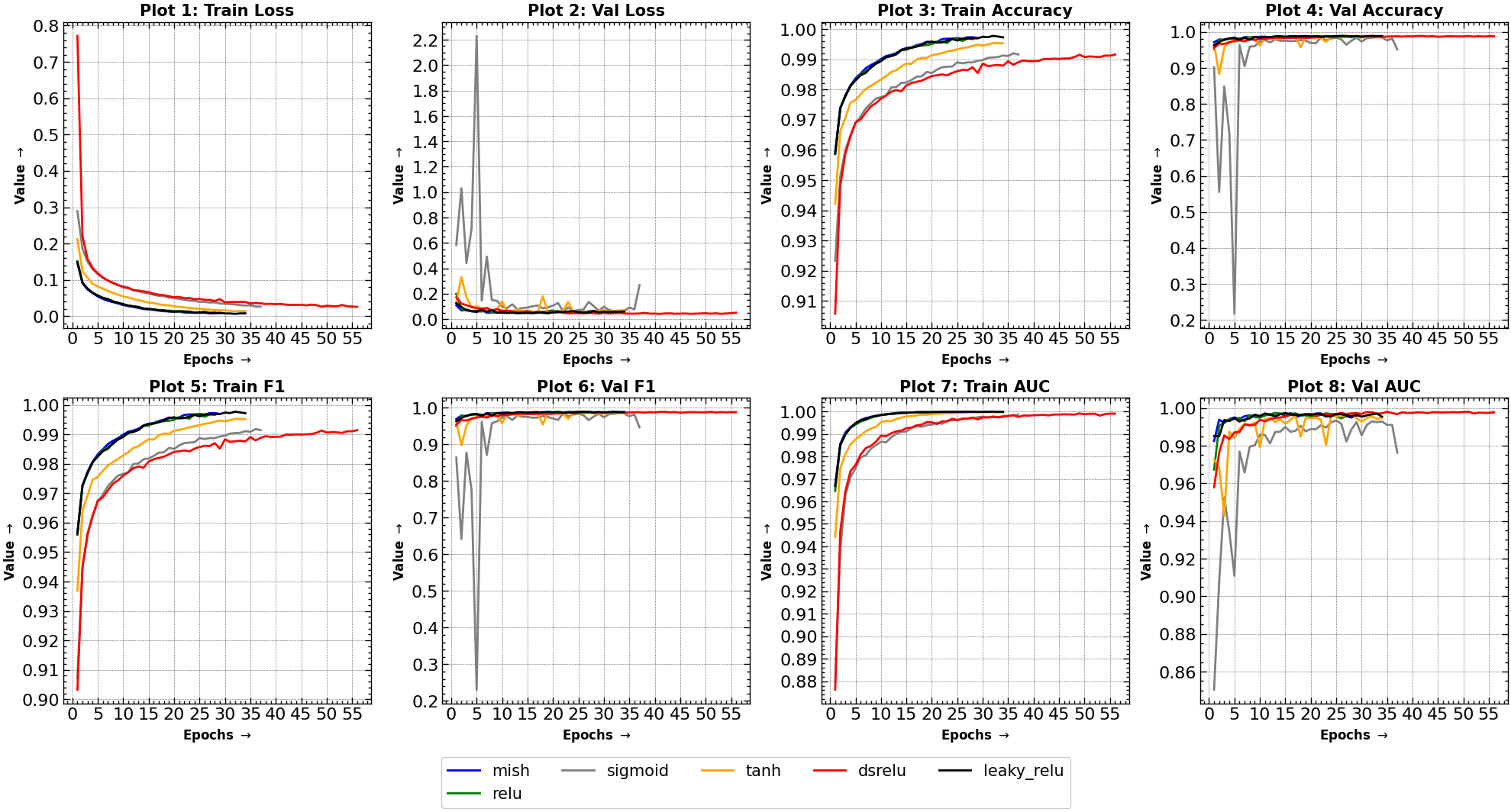}
\caption{Performance of different activation functions on the MIT-BIH Dataset}\label{fig:mitbih_training}
\end{figure}

Refer to Fig \ref{fig:mitbih_training} and Table \ref{mitbih-table} for details. 
As can be seen from the training and validation plots and the performance metrics, DSReLU diminishes the problem overfitting on the F1-score and provides similar performance across all metrics as state-of-the-art activation functions.

\subsection{Training time}

\begin{figure}
\centering
\includegraphics[width=0.4\textwidth]{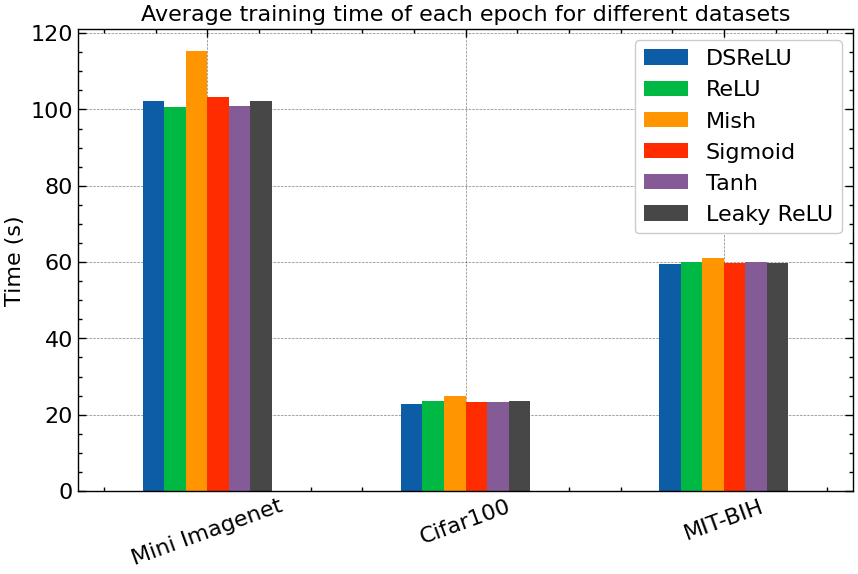}
\caption{Mean training time of ResNet-34 model with different activation functions across different datasets.}\label{fig:time}
\end{figure}

In our study, we used an Early Stopping callback with a patience of 15 epochs to stop training when the model starts overfitting. As can be seen in Fig \ref{fig:imagenet-metrics},\ref{fig:cifar_training},  \ref{fig:mitbih_training}, different models stopped training at different epochs due to this callback. Hence, comparing the total time of training for the different datasets is not possible. Hence, we compare the mean training time for each epoch across the datasets and the activation functions to get an idea of the training speed of DSReLU in comparison with other activation functions across datasets. Please refer to Table \ref{time_comparison} and Fig \ref{fig:time} for details. As is seen from the afore-mentioned table and chart, DSReLU has almost similar (marginally faster) mean epoch training times on the CIFAR-100 and MIT-BIH datasets, and slightly greater time on the Mini ImageNet dataset. The observed performance differences may be attributed to the custom implementation of the DSReLU layer in PyTorch, as compared to the already optimized implementations of traditional activation functions in PyTorch.

\section{Conclusion}

In this study, we introduced a novel dynamic slope variant of the ReLU activation function (DSReLU), designed to optimize learning efficiency and performance in deep neural networks, specifically within the context of computer vision classification tasks. Through systematic experiments employing a ResNet34 architecture, our custom activation function demonstrated improved performance metrics across various datasets, including Mini ImageNet, CIFAR-100, and MIT-BIH datasets, when compared to traditional activation functions such as ReLU, Sigmoid, Tanh, LeakyReLU and even the state-of-the-art Mish activation function.

The dynamic nature of our activation function's slope, which adjusts throughout the training epochs, effectively addresses common issues associated with vanishing gradients and dead neurons, thereby enhancing the model's ability to learn more complex patterns without significant computational overhead. 

Future work will focus on extending the applicability of our dynamic activation function to other deep learning architectures and exploring its effectiveness in a wider range of tasks beyond image classification. Additionally, further investigation into the optimal parameterization of the dynamic slope, including the impact of different initial, final slope values and the steepness of change, will be crucial in maximizing the utility and adaptability of this activation function across diverse learning scenarios.

This research contributes to the ongoing exploration of activation function design, offering a promising avenue for enhancing model performance and efficiency. Our findings underscore the importance of considering activation function adaptability in the pursuit of advancing neural network capabilities, providing a foundation for future innovations in the field.


\bibliographystyle{ieeetr}
\bibliography{main}

\section*{Appendix}

\begin{table*}
\centering
\captionsetup{skip=10pt}
\renewcommand{\arraystretch}{1.5} 
\setlength{\tabcolsep}{8pt} 

\begin{tabular}{|ll|c|c|c|c|c|c|}
  \cline{3-8}
  \multicolumn{2}{c|}{} & \multicolumn{6}{c|}{Activation Function} \\
  \cline{3-8}
  \multicolumn{2}{c|}{} & DSReLU & Mish & ReLU & Sigmoid & Tanh & LeakyReLU \\
  \hline
  \multirow{5}{*}{Train Acc} & 1 & 0.591152 & \textbf{0.980368} & 0.972672 & 0.333284 & 0.953971 & 0.973750 \\
  \cline{2-8}
  & 2 & 0.589069 & \textbf{0.980000} & 0.969730 & 0.331642 & 0.951887 & 0.969706 \\
  \cline{2-8}
  & 3 & 0.587868 & \textbf{0.978382} & 0.964681 & 0.326642 & 0.944265 & 0.967181 \\
  \cline{2-8}
  & 4 & 0.584265 & \textbf{0.976373} & 0.964265 & 0.322132 & 0.938725 & 0.966422 \\
  \cline{2-8}
  & 5 & 0.576201 & \textbf{0.974902} & 0.961618 & 0.317549 & 0.930049 & 0.961103 \\
  \hline
  \multirow{5}{*}{Val Acc} & 1 & \textbf{0.514111} & 0.349333 & 0.373000 & 0.159667 & 0.241222 & 0.420667 \\
  \cline{2-8}
  & 2 & \textbf{0.514444} & 0.402889 & 0.390222 & 0.222222 & 0.229222 & 0.416778 \\
  \cline{2-8}
  & 3 & \textbf{0.485778} & 0.418778 & 0.390111 & 0.139000 & 0.238333 & 0.417889 \\
  \cline{2-8}
  & 4 & \textbf{0.500111} & 0.377778 & 0.397667 & 0.153889 & 0.239667 & 0.392556 \\
  \cline{2-8}
  & 5 & \textbf{0.511000} & 0.406778 & 0.404111 & 0.163333 & 0.261000 & 0.414667 \\
  \hline
  \multirow{5}{*}{Train F1} & 1 & 0.588727 & \textbf{0.980369} & 0.972675 & 0.321360 & 0.953967 & 0.973754 \\
  \cline{2-8}
  & 2 & 0.586834 & \textbf{0.979999} & 0.969725 & 0.319575 & 0.951887 & 0.969703 \\
  \cline{2-8}
  & 3 & 0.585523 & \textbf{0.978382} & 0.964678 & 0.314183 & 0.944258 & 0.967178 \\
  \cline{2-8}
  & 4 & 0.581939 & \textbf{0.976373} & 0.964265 & 0.309418 & 0.938708 & 0.966422 \\
  \cline{2-8}
  & 5 & 0.573714 & \textbf{0.974902} & 0.961608 & 0.304703 & 0.930028 & 0.961097 \\
  \hline
  \multirow{5}{*}{Val F1} & 1 & \textbf{0.510515} & 0.352349 & 0.377149 & 0.140813 & 0.234833 & 0.419223 \\
  \cline{2-8}
  & 2 & \textbf{0.509318} & 0.402392 & 0.394829 & 0.193345 & 0.228613 & 0.419052 \\
  \cline{2-8}
  & 3 & \textbf{0.485347} & 0.418582 & 0.392753 & 0.108845 & 0.239383 & 0.414290 \\
  \cline{2-8}
  & 4 & \textbf{0.492200} & 0.376354 & 0.394762 & 0.130836 & 0.240509 & 0.387551 \\
  \cline{2-8}
  & 5 & \textbf{0.508321} & 0.403009 & 0.403446 & 0.134038 & 0.256055 & 0.413554 \\
  \hline
  \multirow{5}{*}{Train AUC} & 1 & 0.982171 & \textbf{0.999975} & 0.999947 & 0.942067 & 0.999864 & 0.999946 \\
  \cline{2-8}
  & 2 & 0.981722 & \textbf{0.999964} & 0.999937 & 0.940918 & 0.999836 & 0.999938 \\
  \cline{2-8}
  & 3 & 0.981425 & \textbf{0.999964} & 0.999920 & 0.939563 & 0.999802 & 0.999926 \\
  \cline{2-8}
  & 4 & 0.980952 & \textbf{0.999959} & 0.999919 & 0.938372 & 0.999746 & 0.999914 \\
  \cline{2-8}
  & 5 & 0.980508 & \textbf{0.999951} & 0.999902 & 0.936580 & 0.999686 & 0.999903 \\
  \hline
  \multirow{5}{*}{Val AUC} & 1 & \textbf{0.967096} & 0.911652 & 0.921944 & 0.887380 & 0.880892 & 0.941272 \\
  \cline{2-8}
  & 2 & \textbf{0.968000} & 0.932128 & 0.930483 & 0.909604 & 0.872917 & 0.938481 \\
  \cline{2-8}
  & 3 & \textbf{0.962493} & 0.937934 & 0.930360 & 0.873932 & 0.880645 & 0.938755 \\
  \cline{2-8}
  & 4 & \textbf{0.965191} & 0.927474 & 0.935091 & 0.892884 & 0.876501 & 0.932219 \\
  \cline{2-8}
  & 5 & \textbf{0.965986} & 0.934113 & 0.936806 & 0.888145 & 0.886452 & 0.937848 \\
  \hline
\end{tabular} 
\\ 
\caption{Metrics for different Activation Functions on Mini-Imagenet Dataset across 5 phases of 5 fold cross validation. The sequence of rows 1-5 suggest the phases of training. The best metric in each row has been made \textbf{bold}.}
\label{Imagenet-table}
\end{table*}

\begin{table*}
\centering
\renewcommand{\arraystretch}{1.5} 
\captionsetup{skip=10pt}
\setlength{\tabcolsep}{8pt} 

\begin{tabular}{|ll|c|c|c|c|c|c|}
  \cline{3-8}
  \multicolumn{2}{c|}{} & \multicolumn{6}{c|}{Activation Function} \\
  \cline{3-8}
  \multicolumn{2}{c|}{} & DSReLU & Mish & ReLU & Sigmoid & Tanh & LeakyReLU \\
  \hline
  \multirow{5}{*}{Train Acc} & 1 & 0.664375 & \textbf{0.946650} & 0.946300 & 0.234600 & 0.849475 & 0.942325 \\
  \cline{2-8}
  & 2 & 0.652475 & 0.938050 & \textbf{0.943150} & 0.228925 & 0.837175 & 0.933875 \\
  \cline{2-8}
  & 3 & 0.629125 & 0.932175 & \textbf{0.943975} & 0.218500 & 0.825000 & 0.931075 \\
  \cline{2-8}
  & 4 & 0.635150 & 0.927550 & \textbf{0.941450} & 0.209575 & 0.813575 & 0.930150 \\
  \cline{2-8}
  & 5 & 0.613775 & 0.928500 & \textbf{0.939450} & 0.200625 & 0.790350 & 0.922125 \\
  \hline
  \multirow{5}{*}{Val Acc} & 1 & \textbf{0.436200} & 0.399700 & 0.402500 & 0.009600 & 0.354800 & 0.414200 \\
  \cline{2-8}
  & 2 & \textbf{0.461300} & 0.399800 & 0.408900 & 0.009400 & 0.356900 & 0.410400 \\
  \cline{2-8}
  & 3 & \textbf{0.457700} & 0.408500 & 0.408200 & 0.010700 & 0.352300 & 0.414500 \\
  \cline{2-8}
  & 4 & \textbf{0.462400} & 0.411800 & 0.404000 & 0.010500 & 0.341500 & 0.416400 \\
  \cline{2-8}
  & 5 & \textbf{0.459500} & 0.412200 & 0.409700 & 0.009800 & 0.355500 & 0.407700 \\
  \hline
  \multirow{5}{*}{Train F1} & 1 & 0.663391 & \textbf{0.946678} & 0.946293 & 0.213050 & 0.849271 & 0.942313 \\
  \cline{2-8}
  & 2 & 0.651351 & 0.938070 & \textbf{0.943146} & 0.207696 & 0.836956 & 0.933855 \\
  \cline{2-8}
  & 3 & 0.627896 & 0.932210 & \textbf{0.943971} & 0.196406 & 0.824674 & 0.931055 \\
  \cline{2-8}
  & 4 & 0.633925 & 0.927537 & \textbf{0.941445} & 0.187801 & 0.813221 & 0.930150 \\
  \cline{2-8}
  & 5 & 0.612317 & 0.928537 & \textbf{0.939436} & 0.179131 & 0.789847 & 0.922100 \\
  \hline
  \multirow{5}{*}{Val F1} & 1 & \textbf{0.433565} & 0.399366 & 0.401399 & 0.000183 & 0.355555 & 0.414171 \\
  \cline{2-8}
  & 2 & \textbf{0.459378} & 0.398051 & 0.409700 & 0.000175 & 0.355119 & 0.406977 \\
  \cline{2-8}
  & 3 & \textbf{0.456331} & 0.406079 & 0.406335 & 0.000227 & 0.350995 & 0.413718 \\
  \cline{2-8}
  & 4 & \textbf{0.457753} & 0.411667 & 0.402886 & 0.000218 & 0.339138 & 0.414280 \\
  \cline{2-8}
  & 5 & \textbf{0.454617} & 0.410511 & 0.412169 & 0.000190 & 0.356214 & 0.405058 \\
  \hline
  \multirow{5}{*}{Train AUC} & 1 & 0.988405 & 0.999749 & \textbf{0.999820} & 0.908093 & 0.998438 & 0.999770 \\
  \cline{2-8}
  & 2 & 0.987473 & 0.999639 & \textbf{0.999804} & 0.904748 & 0.998141 & 0.999723 \\
  \cline{2-8}
  & 3 & 0.984517 & 0.999598 & \textbf{0.999794} & 0.901412 & 0.997848 & 0.999695 \\
  \cline{2-8}
  & 4 & 0.984238 & 0.999536 & \textbf{0.999769} & 0.897446 & 0.997468 & 0.999667 \\
  \cline{2-8}
  & 5 & 0.983644 & 0.999512 & \textbf{0.999740} & 0.893691 & 0.996853 & 0.999617 \\
  \hline
  \multirow{5}{*}{Val AUC} & 1 & 0.938392 & 0.936380 & 0.938625 & 0.733710 & 0.927083 & \textbf{0.942448} \\
  \cline{2-8}
  & 2 & \textbf{0.948160} & 0.935740 & 0.940575 & 0.782721 & 0.926047 & 0.941256 \\
  \cline{2-8}
  & 3 & \textbf{0.953504} & 0.937808 & 0.938968 & 0.723548 & 0.926261 & 0.942482 \\
  \cline{2-8}
  & 4 & \textbf{0.947334} & 0.940026 & 0.938785 & 0.740547 & 0.923331 & 0.943251 \\
  \cline{2-8}
  & 5 & \textbf{0.952915} & 0.937750 & 0.941620 & 0.745497 & 0.928898 & 0.940885 \\
  \hline
\end{tabular}
\caption{Metrics for Different Activation Functions on CIFAR-100 Dataset across 5 phases of 5 fold cross validation. The sequence of rows 1-5 suggest the phases of training. The best metric in each row has been made \textbf{bold}.}
\label{Cifar-table}
\end{table*}

\begin{table*}
\centering
\captionsetup{skip=10pt}
\renewcommand{\arraystretch}{1.5} 
\setlength{\tabcolsep}{8pt} 

\begin{tabular}{|ll|c|c|c|c|c|c|}
  \cline{3-8}
  \multicolumn{2}{c|}{} & \multicolumn{6}{c|}{Activation Function} \\
  \cline{3-8}
  \multicolumn{2}{c|}{} & DSReLU & Mish & ReLU & Sigmoid & Tanh & LeakyReLU \\
  \hline
  \multirow{5}{*}{Train Acc} & 1 & 0.991448 & 0.997259 & 0.996802 & 0.991548 & 0.995317 & \textbf{0.997716} \\
  \cline{2-8}
  & 2 & 0.991163 & 0.996716 & 0.996616 & 0.991934 & 0.994889 & \textbf{0.997216} \\
  \cline{2-8}
  & 3 & 0.991534 & \textbf{0.996959} & 0.996731 & 0.991034 & 0.995217 & 0.996831 \\
  \cline{2-8}
  & 4 & 0.990806 & 0.996788 & 0.995817 & 0.991220 & 0.994489 & \textbf{0.997259} \\
  \cline{2-8}
  & 5 & 0.990906 & 0.996802 & 0.995346 & 0.990577 & 0.995289 & \textbf{0.997487} \\
  \hline
  \multirow{5}{*}{Val Acc} & 1 & 0.987436 & 0.987893 & 0.988636 & 0.951573 & 0.987208 & \textbf{0.989207} \\
  \cline{2-8}
  & 2 & 0.988293 & 0.988864 & 0.988407 & 0.981783 & 0.985837 & \textbf{0.988693} \\
  \cline{2-8}
  & 3 & 0.987950 & 0.988579 & 0.988636 & 0.976643 & 0.985266 & \textbf{0.989150} \\
  \cline{2-8}
  & 4 & 0.987094 & 0.988065 & 0.988065 & 0.984695 & 0.984638 & \textbf{0.988350} \\
  \cline{2-8}
  & 5 & \textbf{0.988864} & 0.987494 & 0.987950 & 0.984181 & 0.985723 & 0.988522 \\
  \hline
  \multirow{5}{*}{Train F1} & 1 & 0.991333 & 0.997253 & 0.996789 & 0.991440 & 0.995282 & \textbf{0.997712} \\
  \cline{2-8}
  & 2 & 0.991064 & 0.996708 & 0.996604 & 0.991811 & 0.994847 & \textbf{0.997207} \\
  \cline{2-8}
  & 3 & 0.991441 & \textbf{0.996948} & 0.996718 & 0.990900 & 0.995189 & 0.996817 \\
  \cline{2-8}
  & 4 & 0.990691 & 0.996778 & 0.995805 & 0.991107 & 0.994445 & \textbf{0.997251} \\
  \cline{2-8}
  & 5 & 0.990777 & 0.996795 & 0.995321 & 0.990451 & 0.995256 & \textbf{0.997482} \\
  \hline
  \multirow{5}{*}{Val F1} & 1 & 0.987412 & 0.987946 & 0.988526 & 0.946373 & 0.986883 & \textbf{0.989157} \\
  \cline{2-8}
  & 2 & 0.987988 & \textbf{0.988678} & 0.988215 & 0.981557 & 0.985750 & 0.988399 \\
  \cline{2-8}
  & 3 & 0.987633 & 0.988567 & 0.988500 & 0.976680 & 0.984890 & \textbf{0.989056} \\
  \cline{2-8}
  & 4 & 0.986972 & 0.987941 & 0.987834 & 0.984620 & 0.984242 & \textbf{0.988200} \\
  \cline{2-8}
  & 5 & \textbf{0.988687} & 0.987139 & 0.987915 & 0.983861 & 0.985394 & 0.988339 \\
  \hline
  \multirow{5}{*}{Train AUC} & 1 & 0.999177 & 0.999902 & 0.999863 & 0.998583 & 0.999754 & \textbf{0.999927} \\
  \cline{2-8}
  & 2 & 0.999071 & 0.999896 & 0.999857 & 0.998249 & 0.999707 & \textbf{0.999904} \\
  \cline{2-8}
  & 3 & 0.999059 & 0.999887 & 0.999824 & 0.998057 & 0.999697 & \textbf{0.999902} \\
  \cline{2-8}
  & 4 & 0.999020 & 0.999877 & 0.999801 & 0.998001 & 0.999660 & \textbf{0.999884} \\
  \cline{2-8}
  & 5 & 0.999015 & 0.999873 & 0.999782 & 0.997837 & 0.999643 & \textbf{0.999883} \\
  \hline
  \multirow{5}{*}{Val AUC} & 1 & \textbf{0.997706} & 0.994954 & 0.995796 & 0.976228 & 0.995484 & 0.996628 \\
  \cline{2-8}
  & 2 & \textbf{0.997455} & 0.997041 & 0.995383 & 0.990996 & 0.994713 & 0.995390 \\
  \cline{2-8}
  & 3 & \textbf{0.997747} & 0.996076 & 0.996789 & 0.991067 & 0.994991 & 0.996006 \\
  \cline{2-8}
  & 4 & \textbf{0.997133} & 0.995578 & 0.995642 & 0.992865 & 0.992705 & 0.996107 \\
  \cline{2-8}
  & 5 & \textbf{0.998062} & 0.996054 & 0.997248 & 0.992982 & 0.994142 & 0.997054 \\
  \hline
\end{tabular}
\caption{Metrics for Different Activation Functions on MIT-BIH Dataset across 5 phases of 5 fold cross validation. The sequence of rows 1-5 suggest the phases of training. The best metric in each row has been made \textbf{bold}.}
\label{mitbih-table}
\end{table*}

\begin{table}
\centering
\begin{tabular}{l>{\centering\arraybackslash}p{2.5cm}>{\centering\arraybackslash}p{2.5cm}>{\centering\arraybackslash}p{2.5cm}>{\centering\arraybackslash}p{2.5cm}>{\centering\arraybackslash}p{2.5cm}>{\centering\arraybackslash}p{2.5cm}}
\toprule
 & DSReLU & ReLU & Mish & Sigmoid & Tanh & Leaky ReLU \\
\midrule
Mini-ImageNet & 102.092015 & \textbf{100.722127} & 115.313988 & 103.180806 & 100.875571 & 102.197597 \\
Cifar100 & \textbf{22.866360} & 23.609996 & 24.988486 & 23.399378 & 23.346554 & 23.521402 \\
MIT-BIH & \textbf{59.487318} & 59.963315 & 61.142270 & 59.659879 & 59.898150 & 59.783929 \\
\bottomrule
\end{tabular}
\caption{Mean epoch training time for each activation function on different datasets. All values are reported in units of seconds. The least time in each row has been made \textbf{bold}.}
\label{time_comparison}
\end{table}

The source code for the project can be found on GitHub at \href{https://github.com/ScientificArchisman/DSReLU}{ScientificArchisman/DSReLU}.

\end{document}